\documentclass[Afour,sageh,times]{sagej}

\usepackage{moreverb,url}
\usepackage{times}

\usepackage{todonotes}
\usepackage{caption}
\usepackage{subcaption}
\usepackage{soul}
\usepackage{settings}
\usepackage{bm}  % assumes amsmath package installed
\usepackage{comment}
\usepackage{varwidth}
\DeclareCaptionFormat{varwidth}{%
	\begin{varwidth}{\linewidth}#1#2#3\end{varwidth}%
}
\usepackage[colorlinks,bookmarksopen,bookmarksnumbered,citecolor=red,urlcolor=red]{hyperref}

\newcommand\BibTeX{{\rmfamily B\kern-.05em \textsc{i\kern-.025em b}\kern-.08em
T\kern-.1667em\lower.7ex\hbox{E}\kern-.125emX}}

\def\volumeyear{2023}

\newcommand{\robot}{\mathcal{R}}
\newcommand{\domain}{\mathcal{D}}
\newcommand{\task}{\mathcal{T}}
\newcommand{\sourcerobot}{\robot_S}
\newcommand{\sourcedomain}{\domain_S}
\newcommand{\sourcetask}{\task_S}
\newcommand{\targetrobot}{\robot_T}
\newcommand{\targetdomain}{\domain_T}
\newcommand{\targettask}{\task_T}

\begin{document}

\runninghead{Jaquier, Welle, et\,al.}

\title{Transfer Learning in Robotics: \\ An Upcoming Breakthrough? \\ A Review of Promises and Challenges}

\author{No\'emie Jaquier$^\star$\affilnum{1} , Michael C. Welle$^\star$\affilnum{2}, Andrej Gams\affilnum{3}, Kunpeng Yao\affilnum{4}, Bernardo Fichera\affilnum{4}, Aude Billard\affilnum{4}, Ale\v{s} Ude\affilnum{3,5}, Tamim Asfour\affilnum{1}, Danica Kragic\affilnum{2}}

\affiliation{$^\star$ These authors contributed equally (listed in alphabetical order). \\
\affilnum{1}Institute for Anthropomatics and Robotics, Karlsruhe Institute of Technology, Karlsruhe, Germany\\
\affilnum{2}KTH Royal Institute of Technology, Stockholm, Sweden \\
\affilnum{3}Jo\v{z}ef Stefan Institute, Ljubljana, Slovenia\\
\affilnum{4}Learning Algorithms and Systems Laboratory, Ecole Polytechnique F\'ed\'erale de Lausanne, Lausanne, Switzerland\\
\affilnum{5}Faculty of Electrical Engineering, University of Ljubljana, Slovenia\\
}

\corrauth{No\'emie Jaquier, Institute for Anthropomatics and Robotics, Karlsruhe Institute of Technology, Karlsruhe, Germany. \\
Michael C. Welle, KTH Royal Institute of Technology, Stockholm, Sweden. }

\email{noemie.jaquier@kit.edu, mwelle@kth.se}

\begin{abstract}
Transfer learning is a conceptually-enticing paradigm in pursuit of truly intelligent embodied agents. The core concept --- reusing prior knowledge to learn in and from novel situations --- is successfully leveraged by humans to handle novel situations. 
In recent years, transfer learning has received renewed interest from the community from different perspectives, including imitation learning, domain adaptation, and transfer of experience from simulation to the real world, among others. 
In this paper, we unify the concept of transfer learning in robotics and provide the first taxonomy of its kind considering the key concepts of robot, task, and environment. Through a review of the promises and challenges in the field, we identify the need of transferring at different abstraction levels, the need of quantifying the transfer gap and the quality of transfer, as well as the dangers of negative transfer. Via this position paper, we hope to channel the effort of the community towards the most significant roadblocks to realize the full potential of transfer learning in robotics.
\end{abstract}

\keywords{Transfer learning, imitation learning, domain adaptation, sim-to-real, task transfer, embodiment transfer}

\maketitle

\section{The Rise of Transfer Learning in Robotics} 
\label{sec:Intro}

Transferring prior knowledge to novel unknown tasks is one of the abilities that led humans to become the most innovative species on the planet~\citep{Reader16:Innovation}. In particular, humans' capability to transfer cognitive~\citep{Perkins92:TransferOfLearning, Barnett02:FarTransferTaxonomy} and motor skills~\citep{Schmidt87:MotorSkillLearning} from one context to another makes the acquisition of new skills and the resolution of problems possible to a large extent. 
For instance, the difficulty of learning a new language is significantly influenced by factors such as language distance, native language proficiency, and language attitude~\citep{walqui2000contextual} as humans can \textit{transfer} their prior experience, e.g., grammatical constructions or words, from their native language into the new language. In addition, transfer is a key concept in education as the context of learning, e.g., the classroom, significantly differs from the context, e.g., the workplace, where the learned concepts should ultimately be applied~\citep{Perkins92:TransferOfLearning}.

\begin{figure}[tbp]
    \centering
    \includegraphics[width=\linewidth]{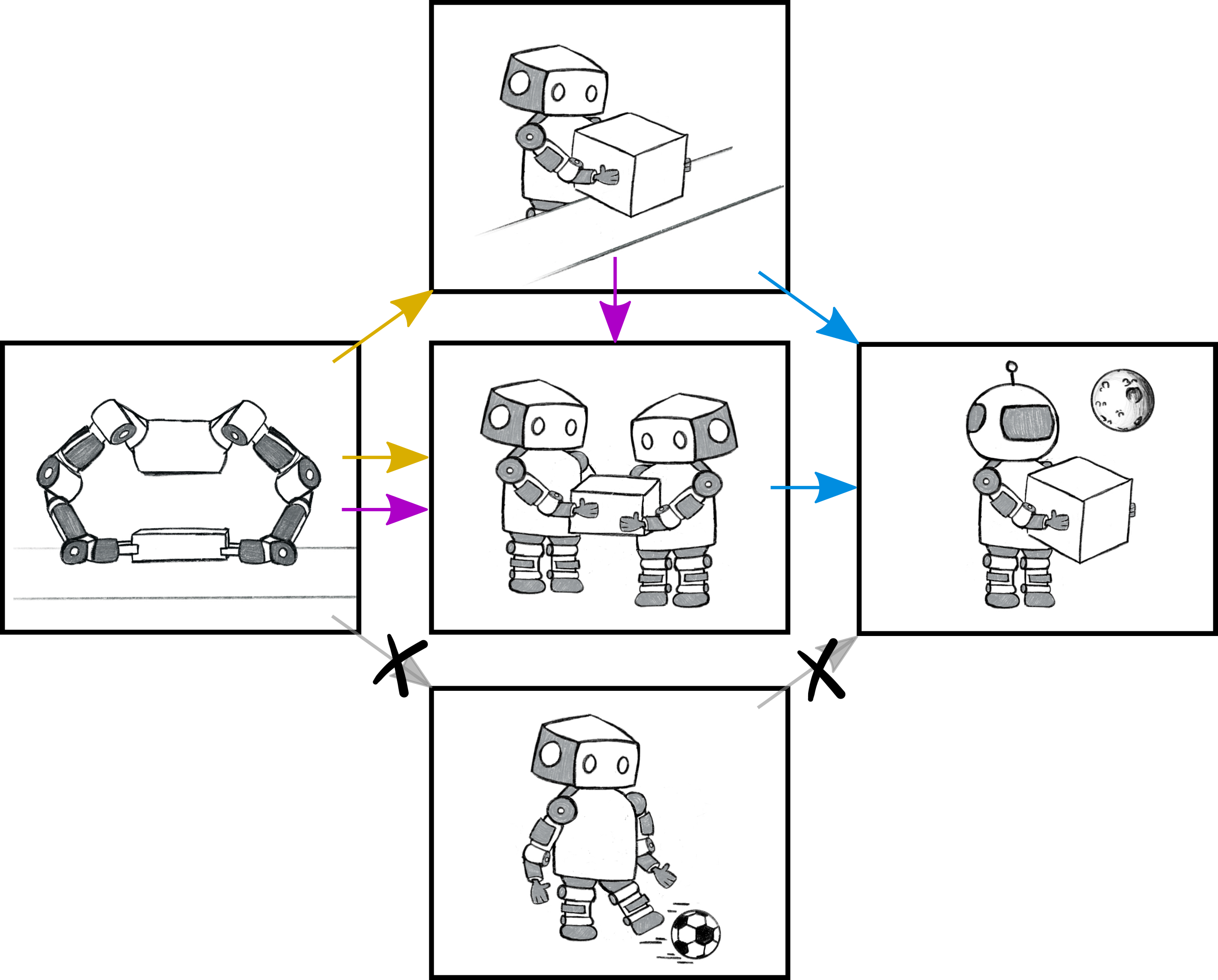}
    \caption{Concept of transfer learning in robotics. The experience of a robot performing a specific task in a specific environment is leveraged to improve the learning of a related task by another robot in a related context. Transfer can occur across embodiments (yellow arrows), across tasks (purple arrows), and/or across environments (blue arrows). It is important to note that successful transfer requires commonalities between the source and target robots, tasks, and environments. For instance, a humanoid robot learning to kick a ball will most likely not benefit from the experience of a dual-arm manipulator systems manipulating a box and vice-versa.}
    \label{fig:ConceptOfTLinRobotics}
\end{figure}

To evolve seamlessly in the real world, robots must feature outstanding cognitive abilities allowing them to perceive their environment, act and react to achieve various goals, and learn continuously from observation and experience, while coping with changes and uncertainty in the world. The transfer learning paradigm for robotics is a promising avenue to avoid learning from scratch by reusing previously-acquired experience in new situations, similar to humans.
The core idea of transfer learning in robotics, illustrated in Figure~\ref{fig:ConceptOfTLinRobotics}, is simple: The experience of a robot performing one task in an environment is leveraged to improve the learning process of a (related) task in a different context, i.e., in a different environment or executed by a different robot.

To identify when transfer learning is needed and/or warranted, one need to identify the similarities and differences between the two situations. Following the concepts proposed in imitation learning~\citep{DautenhahnNehaniv02:ImitationBook}, we distinguish between three aspects that can be deemed similar or different: \emph{The tasks, the environments, and the bodies} (a.k.a. the robots in our case). 

In the conceptual example of Figure~\ref{fig:ConceptOfTLinRobotics}, the experience of the two fixed-based manipulators placing a box on a conveyer belt (left) can be transferred to a humanoid robot executing the same task (middle-top). 
Transfer learning is made possible and easier as the two situations have many aspects in common:
\begin{enumerate}
    \item \textbf{Tasks}: Both tasks are similar as both robots are engaged in moving an object with two arms. The notion of similarity with respect to the object may be relative to the robot's own size and payload. For instance, a pair of industrial arms may be able to support objects up to $20$ kg whereas a humanoid robot may carry only a quarter of this. Yet, the task and strategy to solve it, namely ensuring coordinated movement of both arms and the trajectory to place an object on a conveyor's belt, remains the same. 
     \item \textbf{Environments}: The two environments are similar, as in both cases, the object is to be placed on a conveyor belt. We can also safely assume that, in both cases, the object is subjected to the same external forces (gravity, friction).  
     \item \textbf{Robots}: In this case, the two robots differ. Yet, their differences can be broken into two distinct parts. Both robots are endowed with two arms of similar structure. Solely the second robot is on legs and hence faces the additional challenge of controlling its balance when placing the object on the conveyor belt. This aspect may be resolved without actual learning, for instance, through constrained-based optimization, controlling for additional constraints at the center of mass~\citep{bouyarmane2018quadratic}. Alternatively, reinforcement learning or adaptive control may be used to fine tune gains~\citep{khadivar2023self}.
\end{enumerate}
Hence, in this case, transfer learning is mostly conducted across the different robots, while taking advantage of their similar bimanual structure.
In other cases, when the two robots and/or environments differ importantly, experience may be transferred across tasks. For instance, the bimanual manipulation strategy used when placing a box on a conveyer belt (Figure~\ref{fig:ConceptOfTLinRobotics}, middle-top) may be transferred to a handover task (middle). Such transfer may also be achieved across different tasks and different embodiments: The bimanual strategy of the two fixed-based manipulators (left) may directly be transferred to a different robot executing a related task, e.g., to the humanoid robot handing over the box (middle).
Experience may also be transferred across environments. For example, the bimanual manipulation strategies used by the humanoid robot to place a box on the conveyer belt or to handover the box (Figure~\ref{fig:ConceptOfTLinRobotics}, middle-top and middle) may be reused by a humanoid robot manipulating a box in space (right). In this example, the physical rules of the two environments strongly differ due to the influence of gravity on Earth. 

Finally, in some cases, transferring knowledge may not be beneficial or may even impede the robot performance. 
For instance, we consider transferring experience from the fixed-based bimanual setup placing a box in a conveyer belt to a humanoid robot kicking a ball. In this case, employing transfer learning might not be beneficial. Instead, the large differences between the robots, tasks, and environments might even cause transfer learning to impede the methods employed to fulfill the ball-kicking task. 
It is important to highlight that, even if some particular actions can be transferred, conflicting goals may lead to negative transfer. This clearly showcases the crucial importance of identifying the commonalities and differences between two situations when applying transfer learning in robotics.
Moreover, it highlights the importance of transfer learning metrics that measure not only the transfer quality, but also the transfer gap between robots, tasks, and environments.

\begin{figure}[tbp]
    \includegraphics[width=.9\linewidth]{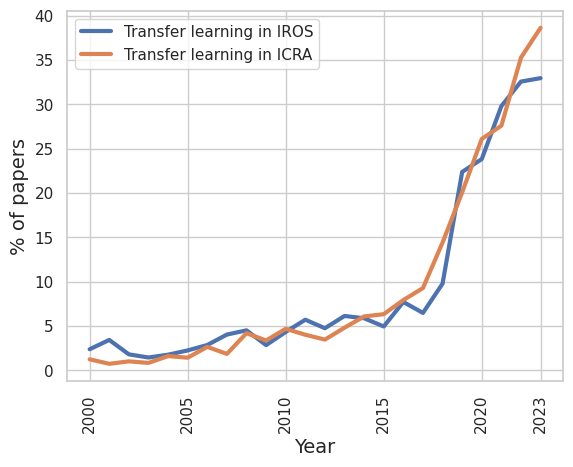}
    \includegraphics[width=.9\linewidth]{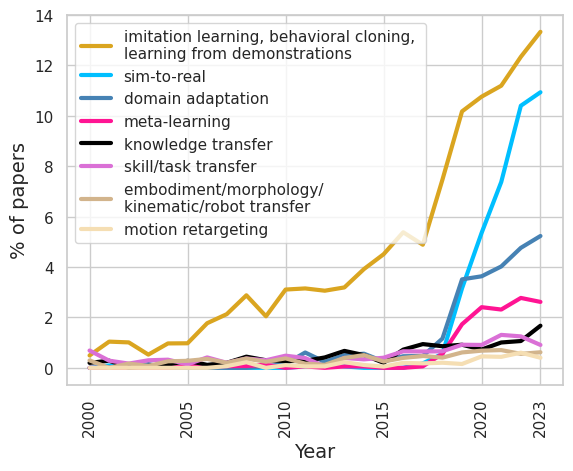}
    \caption{Percentage of papers including words falling under transfer learning in robotics umbrella over the years for the two biggest robotics conferences. The results are based on a systematic search through the content of  44067 papers, excluding their references. \emph{Top:} Percentage per conference. \emph{Bottom:} Percentage per keywords. Keywords related to transfer across embodiments, tasks, and environments are depicted in variations of yellow, purple, and blue. Knowledge transfer is classified independently and thus is depicted in black.}
    \label{fig:overall_trend}
\end{figure}

While the terms falling under the umbrella of transfer learning in robotics are not exactly agreed upon, the robotics community has intensified its effort to transfer various forms of knowledge  across different contexts. Figure~\ref{fig:overall_trend} (top) shows the recent rise in the proportion of published papers at the two biggest robotics conferences --- IROS and ICRA --- invoking keywords\footnote{The considered keywords are: domain adaptation, domain-adaptation, meta-learning, meta learning, sim2real, sim to real, sim-to-real, knowledge transfer, transfer of knowledge, skill transfer, skills transfer, transfer of skills, task transfer, tasks transfer, transfer of tasks, motion retargeting, motion-retargeting, embodiment transfer, transfer embodiments, morphology transfer, transfer morphology, kinematic transfer, transfer kinematic, robot transfer, robot to robot, imitation learning, learning by imitation, imitation-learning, learning from demonstrations, learning from demonstration, learning by demonstrations, learning by demonstration, behavioral cloning.}
that we consider as falling under the umbrella of transfer learning. This recent interest shows that the community strives for embodied transfer learning, which may be a necessary a-priori for truly intelligent systems~\citep{Kremelberg19:Embodiment}.
It is interesting that different keywords related to transfer learning in robotics display different growth rates (see Figure~\ref{fig:overall_trend}, bottom). For instance, terms related to imitation learning, behavioral cloning, and learning from demonstrations display an early rise and a high popularity nowadays, highlighting the effort of the robotics community to tackle transfer between embodiments (human-to-robot or robot-to-robot) from early on~\citep{DautenhahnNehaniv02:ImitationBook}. In contrast, terms related to transfer across environments (i.e., sim-to-real and domain adaptation) only recently gained interest, while terms related to transfer across tasks remain less documented, suggesting that transfer learning across environments and tasks is still in its infancy.

In this position paper, we contend that transfer learning in robotics has the potential to revolutionize the robot learning paradigm by enabling robots to leverage past experience in novel contexts. However, key challenges remain to be addressed to fulfill this potential. In particular, the fundamental question of \emph{identifying the similarities and differences across tasks, environments, and robots in an automatic manner} remains.  
This position paper reviews the successes of the field and identifies relevant research questions and promising directions paving the way forward. Starting from the definition of transfer learning in the machine learning field, we propose a unified definition of transfer learning in robotics and subsequently build a novel taxonomy of transfer learning in robotics based on the key concepts of robot, task, and environment (see Section~\ref{sec:TLtaxonomies}). 
Then, we recount successful applications of transfer learning in robotics and show how they align with the proposed taxonomy (Section~\ref{sec:SuccessesTLinRobotics}). In Section~\ref{sec:ChallengesAndPromisingDirections}, we outline challenges, as well as promising research directions to tackle them, including abstraction levels and universal representations for transfer learning in robotics, interpretability, benchmarks and simulations, transfer learning metrics, and dangers of negative transfer. Last but not least, we call for actions to address the most immediate roadblocks in Section~\ref{sec:Conclusion}.
Overall, the contributions of this paper are twofold: \emph{(1)} We provide a unified view of transfer learning in robotics by comprehensibly defining the notion of transfer learning in robotics and by introducing the first taxonomy of transfer learning in robotics; \emph{(2)} We provide a review of promises and challenges in the field of transfer learning in robotics, identifying the most significant roadblocks on the way to unraveling the full potential of transfer learning in robotics.

\section{Transfer Learning Taxonomies: From Machine Learning to Robotics}
\label{sec:TLtaxonomies}
While the machine learning community has devoted substantial efforts to defining and systematizing transfer learning and categorizing its different instances, transfer learning in robotics is found under various terminologies. This section aims at providing a unified view of transfer learning in robotics. To do so, we take inspiration from machine learning taxonomies and define a taxonomy of transfer learning settings that occur in robotics.

\subsection{Taxonomy of Transfer Learning in Machine Learning}

\begin{figure}[tbp]
    \centering
    \includegraphics[width=\linewidth]{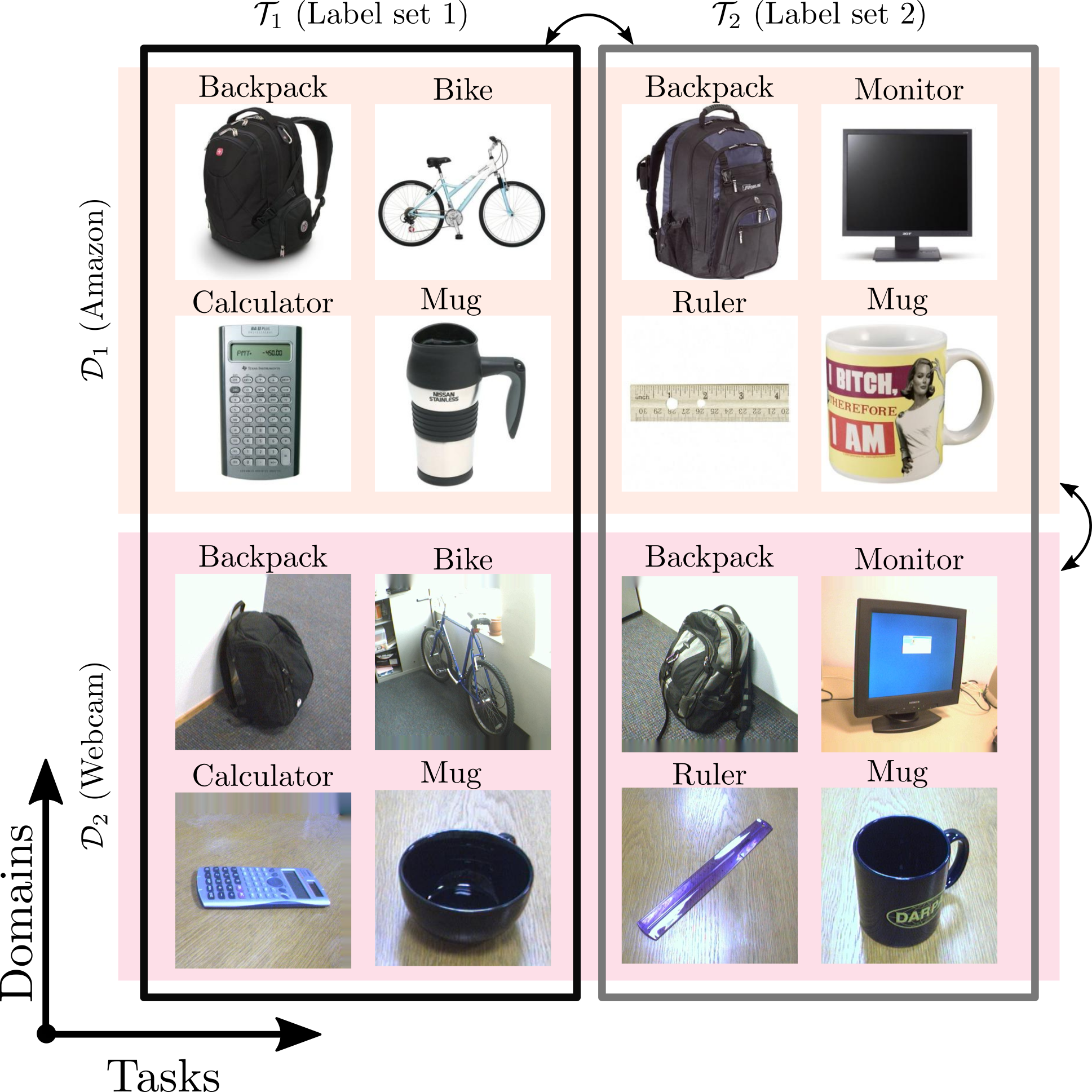}
    \caption{Example of transfer learning on the Office 31 dataset~\citep{saenko2010adapting}. Transfer can occur \emph{(1)} between the two label sets (tasks $\task_1$ and $\task_2$), \emph{(2)} between the two sources used to obtain the images (domains $\domain_1$ and $\domain_2$), and \emph{(3)} between both tasks and domains. The different transfer learning instances are illustrated with black arrows.}
    \label{fig:ml_tl_overview}
\end{figure}

In this section, we introduce the definitions that are commonly adopted in the transfer learning community, see e.g.,~\citep{pan2010survey,zhuang2020comprehensive,yang_zhang_dai_pan_2020:TL}. 
Transfer learning builds on the two fundamental concepts of domain and task. 
A \emph{domain} $\domain = \lbrace \mathcal{X}, p(X) \rbrace$ consists of a feature space $\mathcal{X}$ and a marginal distribution $p(X)$, with $X$ denoting an instance set of the feature space such that $X = \lbrace \bm{x}_i \rbrace_{i=1}^N$, $\bm{x}_i\in \mathcal{X}$.
Given a specific domain $\domain$, a \emph{task} $\task = \lbrace \mathcal{Y}, f \rbrace$ consists of a label space $\mathcal{Y}$ and a predictive function $f: \mathcal{X} \to \mathcal{Y}$. The predictive function is used to predict new labels $\bm{y}\in\mathcal{Y}$ associated with a new instance $\bm{x}\in\mathcal{X}$. It is typically learned from a training dataset $\lbrace \bm{x}_i, \bm{y}_i \rbrace_{i=1}^M$, with $\bm{x}_i \in \mathcal{X}$, $\bm{y}_i \in \mathcal{Y}$. 
While standard machine learning approaches assume that training and test datasets share common domains and tasks, in the case of transfer learning they may instead belong to different spaces, referred to as \emph{source} and \emph{target} spaces. Therefore, it also has the potential to tackle open-set problems~\citep{Geng21:OpenSetRecognition}. An example from computer vision is shown in Figure~\ref{fig:ml_tl_overview}.

Transfer learning approaches are commonly categorized into inductive, transductive, and unsupervised transfer learning~\citep{pan2010survey,zhuang2020comprehensive,yang_zhang_dai_pan_2020:TL}. This categorization focuses on the availability of labels independently of the relationships between source and target spaces. Namely, in the inductive setting, labels are available in both source and target spaces, while they are only available in the source space in the transductive setting, and are not available in any space in the unsupervised setting. 
For a more encompassing categorization that generalizes to robotics, we instead propose to focus on the fundamental concepts of domain and task and on their relationship in the source and target spaces.

\begin{definition}[Transfer Learning in Machine Learning]
\label{def:TL_ML}
Let $S = \lbrace \sourcedomain, \sourcetask \rbrace$ a source space and $T = \lbrace \targetdomain, \targettask \rbrace$ a target space. The objective of \emph{transfer learning} is to improve the learning of the predictive function $f_T$ over the target domain $\targetdomain$ by taking advantage of knowledge from the source domain $\sourcedomain$, and task $\sourcetask$, where $ \sourcedomain \neq  \targetdomain$ and / or $\sourcetask \neq  \targettask$.
\end{definition}

\begin{figure*}[tbp]
    \centering
    \includegraphics[width=.7\linewidth]{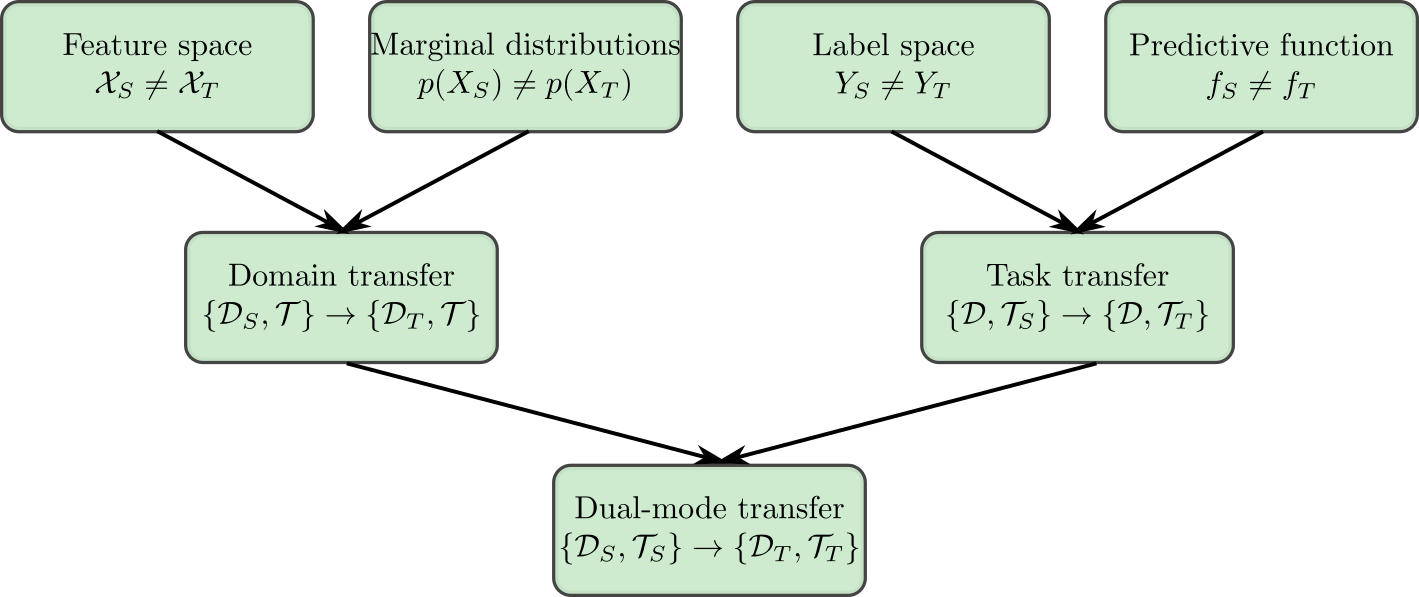}
    \caption{Hierarchical taxonomy of transfer learning in the context of machine learning. Domain transfer occurs when the source and target domain differ and is characterized by a difference in the feature space and/or in the marginal distribution. Task transfer occurs when the source and target task differ and is characterized by a difference in the label space and/or in the predictive function. Dual-mode transfer occurs when both the source and target domain and task differ.
    }
    \label{fig:TlinML-categorization}
\end{figure*}

We observe that Definition~\ref{def:TL_ML} implies the following hierarchical taxonomy of transfer learning settings illustrated in Figure~\ref{fig:TlinML-categorization}.
Note that source space may correspond to a union of multiple sub-source tasks and/or domains.
\begin{enumerate}
\item \textbf{Task transfer learning}: $\lbrace  \domain, \sourcetask \rbrace \to \lbrace  \domain, \targettask \rbrace$.
In this setting, the target task differs from the source task, $\sourcetask \neq \targettask$. This can indicate a difference in the label space, the predictive function or both. Notice that we refer to task transfer learning whenever the source and target domains are identical while the source and target tasks differ. The extent and conditions of the task differences were further categorized according to various transfer learning taxonomies, see, e.g.,~\citep{pan2010survey,zhuang2020comprehensive,yang_zhang_dai_pan_2020:TL}.

Task transfer learning approaches include learning strategies involving Gaussian process (GP) prior sharing across different tasks~\citep{lawrence2004learning,bonilla2007multi}. 
Other strategies focus on sharing the parameters of the model itself rather than the hyperparameters. One important category of algorithms in this domain takes advantage of a modified version of support vector machine (SVM) to transfer knowledge between source and target spaces~\citep{evgeniou2004regularized,li2012cross}. In this modified SVM, the model's parameters consist of a part shared across the source and target spaces, while the other part is space-specific. The uniqueness of the solution (learning efficiency) and model interpretation make these convex optimization algorithms an interesting solution for robotics.
Multilinear relationship networks~\citep{long2017learning} leverages labeled data from related source domains by adopting a Bayesian framework for the task-specific portion of the network.

\item \textbf{Domain transfer learning}: $\lbrace  \sourcedomain, \task \rbrace \to \lbrace  \targetdomain, \task \rbrace$.
This form of transfer learning occurs when the source and target tasks are identical, $\sourcetask = \targettask$, but the source and target domains differ. The condition $\sourcedomain \neq \targetdomain$ can indicate a difference in the feature space, in the marginal distribution, or in both. It is generally assumed that the domains are related to a certain extend.

The difference in marginal distribution is often tackled by learning a mapping between overlapping instances (also known as ``support'') between the source and target domains $\sourcedomain$ and $\targetdomain$.
Such approaches primarily rely on instance weighting strategies, such as assigning weights to instances or labeled data in $\sourcedomain$ for reuse in $\sourcedomain$. For instance, kernel mean matching~\citep{huang2006correcting} matches source and target domain instance means within a reproducing kernel Hilbert space.
In the case of different feature spaces ($\mathcal{X}_S \neq \mathcal{X}_T$), existing approaches aim at reducing
domain differences while preserving the properties or structures within the same domain.
For instance, structural correspondence learning~\citep{blitzer2006domain} utilizes pivot features to establish pseudo tasks connected to the target task and applies multi-task learning techniques to model relationships between pivot features and other features. Spectral feature alignment~\citep{pan2010cross} models inter-dependencies between pivot features and other features using a bipartite graph and identifies novel common features through spectral clustering methods applied to the graph.

\item \textbf{Dual-mode transfer learning}: $\lbrace  \sourcedomain, \sourcetask \rbrace \to \lbrace  \targetdomain, \targettask \rbrace$.
In this scenario, both the source and target domains and tasks differ. This is the most challenging setting in transfer learning as every additional difference between the source and target label space, predictive function, feature space, and marginal distribution increases the complexity of the problem.

Approaches in the dual-mode are currently mostly related to unsupervised transfer learning scenarios. This remains an underexplored area due to the difficulty of capturing the similarities --- or the transferable information (instance, feature, parameter, etc.) --- between the source and target spaces.

\end{enumerate}

Figure~\ref{fig:ml_tl_overview} illustrates the aforementioned transfer learning setting using the transfer dataset Office 31~\citep{saenko2010adapting}, a classical benchmark for transfer learning. In this case, the domains correspond to the source used to obtain the images (i.e., Amazon and a webcam in Figure~\ref{fig:ml_tl_overview}) and the tasks correspond to the labels of the objects represented in the images.

To extend these concepts to robotics, we must consider the robot as an additional mode. In the next section, we discuss its implications for transfer learning in robotics.

\subsection{Taxonomy of Transfer Learning in Robotics} 

\begin{figure*}[tbp]
    \centering
    \includegraphics[width=.8\linewidth]{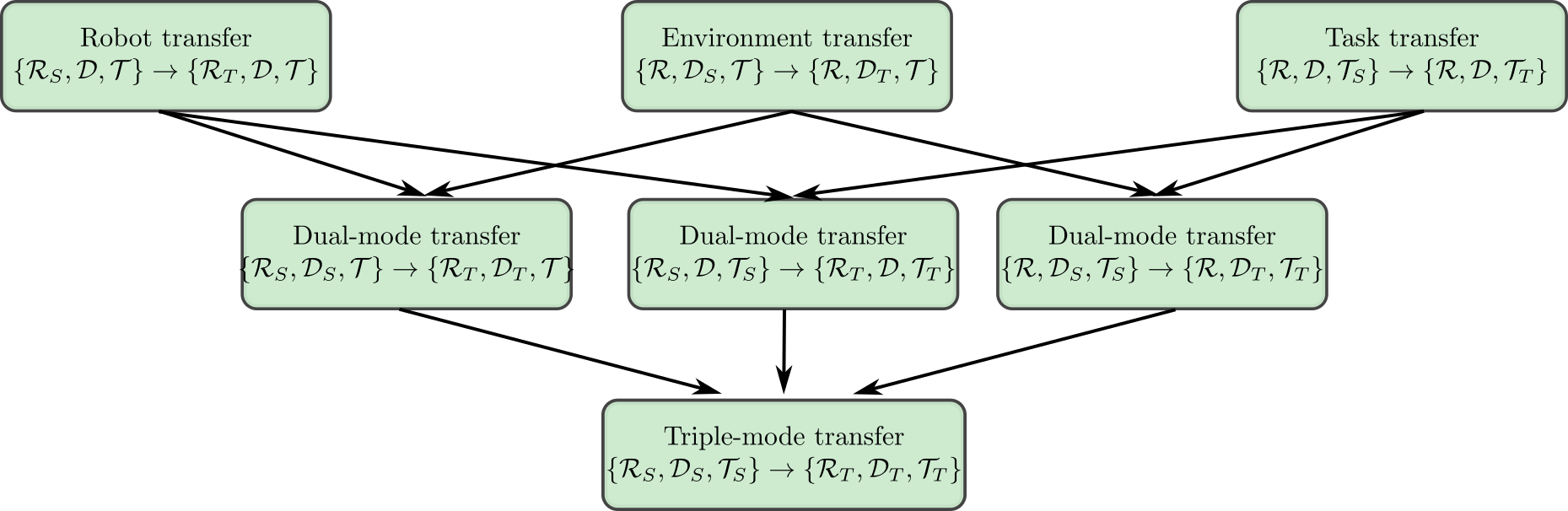}
    \caption{Hierarchical taxonomy of transfer learning in the context of robotics. Robot, environment, and task transfer occur when the source and target robot, environment, and task differ, respectively. Dual-mode and triple-mode transfer occur when two of these modalities and the three of them differ, respectively.}
    \label{fig:TlinR-categorization}
\end{figure*}

Transfer learning in robotics builds on the three fundamental concepts of robot, environment, and task. A \emph{robot} $\robot$ is defined as an embodiment that can act in and thus influence its environment. It encompasses a body with defined morphology, kinematics, dynamics, and sensor modalities. In robotics, the domain $\domain$ is generally considered equivalent to the \emph{environment}, which is defined as the virtual or physical world in which the robot lives and interacts. 
The robot accesses the state of the environment via sensory observations, e.g., images, contact forces, auditory and olfactory signals. 
Informally, the \emph{task} $\task$ refers to what the robot is required to do in the environment. More formally, a task is a discrete or continuous (sub)goal that can be achieved by the robot through (inter)actions within the environment.
In general, the goal of the robot is to perform a given task in the environment. The goal of transfer learning in robotics is to leverage prior knowledge from a source space, composed by a robot, a task, and an environment, to improve the performance in a target space, where one or more mode differs from the source space.
Formally, we define transfer learning in robotics as an analogy of the machine learning definition~\ref{def:TL_ML} as follows.
\begin{definition}[Transfer Learning in Robotics]
\label{def:TL_robotics}
Let $S = \lbrace \sourcerobot, \sourcedomain, \sourcetask \rbrace$ a source space and $T = \lbrace \targetrobot, \targetdomain, \targettask \rbrace$ a target space. The objective of \emph{transfer learning} in robotics is to improve the performance of the robot $\targetrobot$ executing the task $\targettask$ in the environment $\targetdomain$ by taking advantage of knowledge from the source robot\footnote{Note that the source $\sourcerobot$ may be a human instead of a robot. This typically occurs, e.g., in imitation learning.} $\sourcerobot$, environment $\sourcedomain$, and task $\sourcetask$, where at least one element of the target space $T$ is different from its counterpart in the source space $S$.
\end{definition}
It is important to emphasize that, unlike transfer learning in machine learning which only involves disembodied agents, the agent's \emph{embodiment} --- in other words, the \emph{robot} --- is key for transfer learning in robotics. This introduces additional challenges: The presence of a robot not only adds an additional mode to the transfer learning problem and thus to the hierarchical categorization, but also brings numerous robotics-specific issues. Transfer learning methods for robotics must cope with the fact that robots are embodied agents that act and interact in the real world.

\begin{figure*}[tbp]
     \centering
     \includegraphics[width=.8\linewidth]{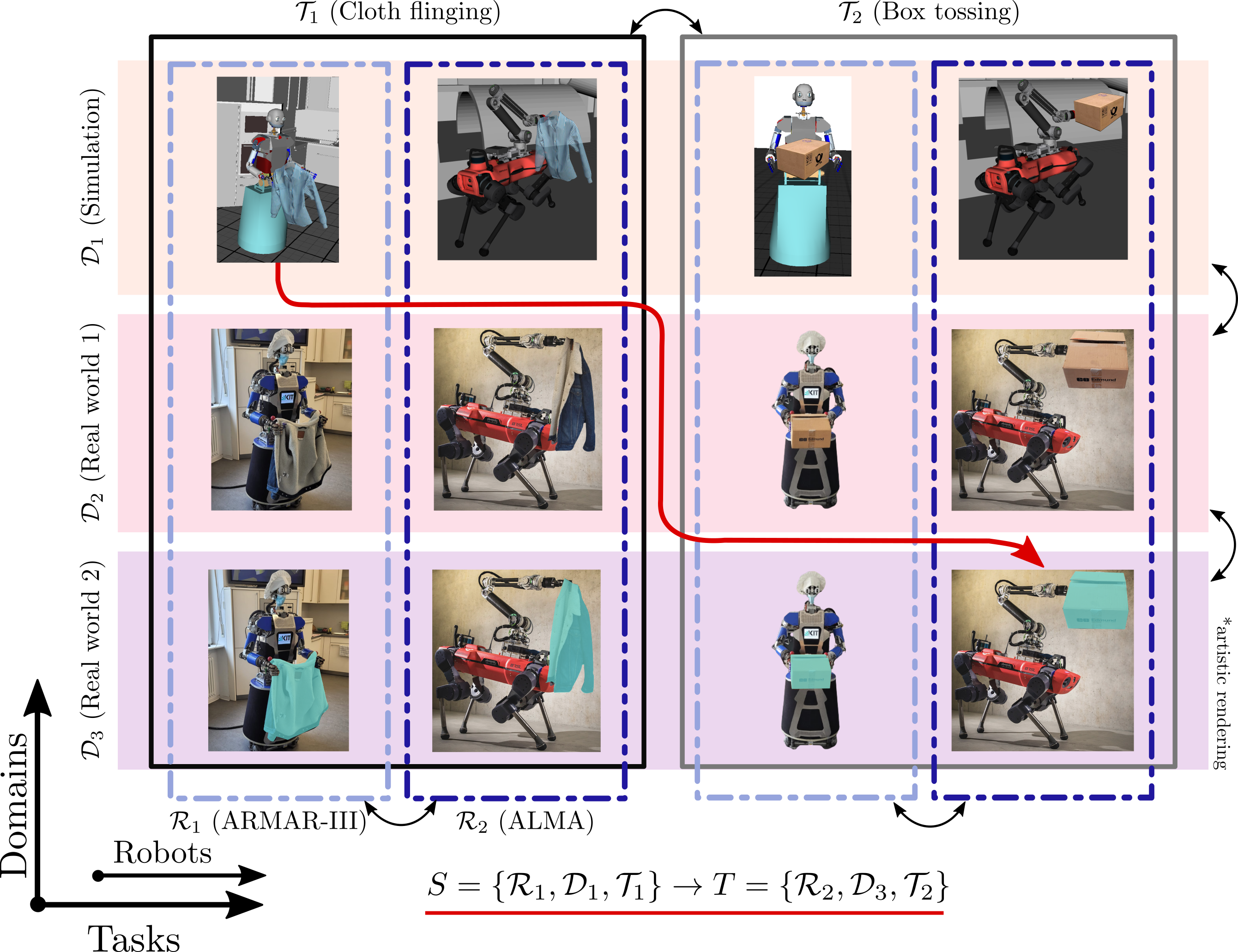}
     \caption{Illustration of the categories of the hierarchical taxonomy for transfer learning in robotics. The humanoid dual-arm robot ARMAR-III ($\robot_1$) and the four-legged robot ALMA equipped with a manipulator ($\robot_2$) execute a cloth flinging task ($\task_1$) and a box tossing task ($\task_2$) in three different environments, namely a simulator ($\domain_1$), and in the real world with different object instances ($\domain_2$ and $\domain_3$). Transfer learning can occur \emph{(1)} between the two robots, \emph{(2)} between two environments, \emph{(3)} between the two tasks,  and \emph{(4-5)} between two or three instances thereof. The different transfer learning instances are illustrated with black arrows. Triple-mode transfer learning, which reuses knowledge from a source space to a target space with different robots, environments, and tasks, is depicted by a red arrow.}
     \label{fig:TL_categories_robotics}
\end{figure*}

Inspired by the hierarchical taxonomy defined for transfer learning in machine learning community, we propose a hierarchical taxonomy for transfer learning in robotics based on the relationship between the source and target robots, environments, and tasks, as shown in Figure~\ref{fig:TlinR-categorization}. Specific illustrative examples of its categories are depicted in Figure~\ref{fig:TL_categories_robotics}.
Our taxonomy considers the following settings:
\begin{enumerate}
    \item \textbf{Robot transfer learning}: $\lbrace \sourcerobot, \domain, \task \rbrace \to \lbrace \targetrobot, \domain, \task \rbrace$. The goal of this setting is to endow a target robot with the ability to perform a given task known by other source robot(s) in the same environment. Note that the source and target robots may have (very) different morphologies, kinematics, and sensor modalities, leading to different capabilities. For example, Figure~\ref{fig:TL_categories_robotics} illustrates a transfer between the humanoid dual-arm robot ARMAR-III~\citep{Asfour2006:ARMARIII} and ALMA~\citep{8794273}, a four-legged robot equipped with a robotic arm. 
    Moreover, the transfer can happen at different actions levels, e.g., at the level of joint or task-space controllers, or at the planning level, and at different perceptual levels, e.g., across different sensors~\citep{Tatiya20:HapticKnowledgeTransfer} or sensory modalities~\citep{lee2020making}.
    Instances of robot transfer learning are \emph{(i)} imitation learning~\citep{SCHAAL1999}, where a teacher human or robot provides demonstrations of a task to a student robot that learns to reproduce the given task in the same environment, \emph{(ii)} (goal-directed) motion retargeting~\citep{Dariush08:MotionRetargeting,Yin23:DanceStyleTransfer}, whose goal is to learn a mapping between different kinematic structures, and
    \emph{(iii)} perceptual transfer, where the robots are equipped with different sensory modalities~\citep{Silva20:PlayingGamesDark} such as touch, vision, sound, or olfaction. 
    \item \textbf{Environment transfer learning}: $\lbrace \robot, \sourcedomain, \task \rbrace \to \lbrace \robot, \targetdomain, \task \rbrace$. This setting aims at transferring the ability of a robot to perform a given task in a source environment to a different target environment. Its main challenge is to overcome the mismatch between source and target environments in terms of data and environment parameters such as, e.g., underlying dynamics or transition models. For instance, models learned for a specific task performed on earth typically needs to be adapted to perform the same task in an underwater environment or in space. This requires identifying which physical parameters differ between $\sourcedomain$ (the earth) and $\targetdomain$ (the underwater environment, or space).
    Typical instances of environment transfer learning are \emph{(i)} domain adaptation~\citep{Bousmalis18:DomainAdaptationGrasping, Wang21:DomainAdaptation}, and \emph{(ii)} sim-to-real transfer~\citep{Muratore2022}. The latter is a particular case of the former in which the experience is explicitly transferred from a simulation environment --- in which training data are inexpensive and models are fast to train --- onto the real world.
    Sim-to-real transfer is showcased by the first and second rows in Figure~\ref{fig:TL_categories_robotics}. The second and third rows indicate transfer between two real-world environments, where the objects composing the physical environment (the cloth or the box) differ. 
    \item \textbf{Task transfer learning}: $\lbrace \robot, \domain, \sourcetask \rbrace \to \lbrace \robot, \domain, \targettask \rbrace$. This setting aims at leveraging the ability of a robot to perform a given task to learn how to execute a different task in the same environment. The underlying assumption is that the source and target tasks are --- to some extent --- similar, so that experience can be reused between source and target tasks. For instance, the box tossing and cloth flinging tasks of Figure~\ref{fig:TL_categories_robotics} share similar dynamics characteristics: In both cases, the robot must generate high-velocity dynamic actions to successfully execute the task. Therefore, we may expect that experience about box tossing may be reused by the robot when learning to fling a cloth. Challenges of task transfer learning include inferring which part of the source task experience should be transferred and at which level (joint or task space, planning, etc).
    Notice that generalizing a given task to an unseen context is a special case of task transfer learning~\citep{Mandlekar20:GTI,Li23:TaskGeneralization}. In this case, the model is made compatible with different instances of the same task. 
    Curriculum learning~\citep{narvekar2020curriculum,Shukla22:ACuTe} is also a special case of task transfer learning, where a sequence of intermediary tasks of gradually-increasing difficulty is used to learn a complex target task.
    Finally, task transfer is also particularly considered in the areas of lifelong learning, where task transfer is considered based on a never-ending stream of data, and compositional learning, which focuses on transfer across compositionally-related tasks~\citep{mendez2022reuse,mendez2021lifelong}.
    \item \textbf{Dual-mode transfer learning}: $\lbrace \sourcerobot, \sourcedomain, \task \rbrace \to \lbrace \targetrobot, \targetdomain, \task \rbrace$, $\lbrace \sourcerobot, \domain, \sourcetask \rbrace \to \lbrace \targetrobot, \domain, \targettask \rbrace$, or $\lbrace \robot, \sourcedomain, \sourcetask \rbrace \to \lbrace \robot, \targetdomain, \targettask \rbrace$. This setting is concerned by transferring knowledge between two spaces which differ across two modes. It assumes that the similarities between source and target spaces can still be leveraged when they share a single common mode. For instance, in Figure~\ref{fig:TL_categories_robotics}, it is reasonable to assume that experience acquired in simulation to toss a box may be reused to fling a cloth with the same robot in the real world. Dual-mode transfer learning remains largely unexplored in robotics due to the additional level of complexity compared to the single-mode transfer setting listed above, which has not yet been fully resolved.
    \item \textbf{Triple-mode transfer learning}: $\lbrace \sourcerobot, \sourcedomain, \sourcetask \rbrace \to \lbrace \targetrobot, \targetdomain, \targettask \rbrace$. This setting assumes that all three modes of the source and target spaces differ. It is inspired by the human ability to successfully acquire knowledge by observing others executing similar tasks in different environments. For instance, one may observe a chef cooking a pie in a restaurant kitchen and reuse some of her techniques to cook a cake in her own non-professional kitchen. Reusing knowledge from a source space in a target space with different robots, environments, and tasks would endow robots with human-like generalization abilities. This setting is the most challenging, and requires bridging the gaps between high-level semantic information --- indicating the degree of similarity between spaces --- and low-level actions. It is the ultimate goal of transfer learning in robotics, as indicated by the red arrow in Figure~\ref{fig:TL_categories_robotics}.
\end{enumerate}

Notice that, depending on the relationship between the source and target spaces, our Definition~\ref{def:TL_robotics} intrinsically refers to related fields, some of which received significant attention over the years. In Figure~\ref{fig:overall_trend} (bottom), we notably observe that imitation learning is the most mentioned transfer learning field followed by sim-to-real and domain adaptation.
In this sense, we view transfer learning in robotics as an umbrella term that encompasses ``imitation learning", ``learning from demonstrations", ``sim-to-real", ``domain adaption", ``meta-learning", ``knowledge transfer", ``skill transfer", ``motion retargeting", ``embodiment transfer", ``morphology transfer", and ``kinematic transfer", among others.

\section{Successes of Transfer Learning in Robotics}
\label{sec:SuccessesTLinRobotics}
Change of environment or domain as in $\lbrace \robot, \sourcedomain, \task \rbrace \to \lbrace \robot, \targetdomain, \task \rbrace$, change of task as in $\lbrace \robot, \domain, \sourcetask \rbrace \to \lbrace \robot, \domain, \targettask \rbrace$ and change of the robot as in $\lbrace \sourcerobot, \domain, \task \rbrace \to \lbrace \targetrobot, \domain, \task \rbrace$ have all been addressed with varying success in transfer learning in robotics. The body of literature on the topic is extremely vast, making a comprehensive overview beyond the scope of this paper. On the other hand, research activities that by definition fit into the scope of transfer learning have been addressed before the term took root in robotics. An example of such is imitation learning~\citep{SCHAAL1999}, where task execution knowledge is transferred from the human to the robot, or generalization, where task execution knowledge is transferred to (at least) a variation of the task~\citep{Ude2010}. In the following, we provide examples of transfer learning in robotics, also in the light of such above-mentioned applications.

\subsection{Environment Transfer}
Change of environment conditions~\citep{Kramberger2016} or the complexity of the environment where the task is being executed~\citep{vosylius2022} provide examples of generalization to a declaratively new environment. However, the environment (domain), can be different in other aspects that go beyond just the setting -- e.g. contact conditions or other physical conditions might not be the same~\citep{Muratore2022}. One example of such is transfer from the simulation-to-reality or sim-to-real. 

Potentially unjustly, but transfer learning in robotics is often associated exactly with sim-to-real, where typically experience is obtained in one domain --- the simulation ---, and exploited to accelerate learning in the transferred domain --- the real world. Several reviews cover sim-to-real transfer learning in robotics, i.\,e.,~\citep{Muratore2022, Zhao2020}, affirming the notion of a huge body of work in this field. The gist of sim-to-real lies in  the notion that collecting the data for modern (deep) learning and other AI algorithms in the real world is too expensive in terms of time and resources to scale up~\citep{Muratore2022}. Therefore, the data is collected in simulation, despite the difference between the real and simulated domains. This difference, referred to as the ``reality gap"~\citep{Collins2019}, needs to be overcome for real world execution, which is done using transfer learning. Since collecting data in the real world is so time-consuming and expensive, researchers might change the domain to a different simulation, ending up with sim-to-sim methodologies. These are applied to demonstrate the behavior of transfer learning methodologies. 

Different practices have been proposed for sim-to-real transfer learning, starting with realistic modelling~\citep{Muratore2022}. No matter how accurate, modelling will never be fully cover all the aspects of the real world~\citep{Muratore2022}, thus other approaches have emerged. Domain randomization, such as randomization of image backgrounds, of physical parameters of objects and robot actions, or of controller parameters~\citep{Hofer2021}, is a common approach. By randomizing over, for example, physical parameters, the approach tries to cover the entire spectrum of these parameters in the hope that this includes the parameters that describe the real world. Even so, one-shot transfer learning is seldom successful~\citep{Zhao2020}, and additional learning is required, for example using reinforcement learning~\citep{ada2022}, back-propagation~\citep{Chen2018} or both~\citep{Loncarevic2022}. If there are significantly fewer learning iterations in the target domain, the process is called few-shot transfer learning~\citep{Ghad2021}. Few-shot transfer learning has notably been applied to sim-to-real transfer in robotics~\citep{bharadhwaj2019data, shukla2023framework}.
Given that more information can be available in the simulation, the notion of privileged learning was introduced, where the privileged information is used to train a high performance policy, which in turn trains a proprioceptive-only student policy~\citep{Hutter2020}. The idea was very successfully demonstrated in quadrupedal locomotion by more than one group~\citep{Hutter2020, Kumar2022}, and is general enough to be applied for very different tasks, such as excavator walking~\citep{Egli2022} and even robotized handling of textiles~\citep{longhini2022edo}. 

\subsection{Task Transfer}
Transferring of robot walking from one domain to the other can be considered more than just domain transfer, as walking itself can be different for different environments. For instance, a pacing gait learned to walk on smooth ground might not be stable enough for walking on mountainous terrains. Thus, walking on mountainous terrains can be seen as a novel task, which may benefit from transferring previously-learned gaits adapted to other terrains.
Moreover, walking is not an isolated instance: If the robot can learn to throw accurately at one target, a modulation of the throwing task to aim at a different target can in fact be considered at the least a different instance of the same task, if not a different task overall. 

Such transfers from one (or several) task instances to a new one have been utilized in robotics before, and were often referred to as generalization. In this sense, for example, fast learning from a small set of demonstrations was applied with nonlinear autonomous dynamical system (DS), which have the ability to generalize motions to unseen contexts~\citep{Khansari2011}. Similarly, a set of dynamical systems in the form of dynamic movement primitives was used to generalize to transfer knowledge from known situations to unknown in positions~\citep{Ude2010} and in torques~\citep{Denisa2016}, probabilistic movement primitives (ProMPs) encode complete families of motions~\citep{Paraschos2013}, TP-GMMs adapt to changes of predefined local frames~\citep{Calinon2018}, and Mixture Density Networks adapt a learned motion primitive to new targets specified in a different space~\citep{Zhou2020}\footnote{Note that demonstrations and reproductions performed with the same robot avoid the need for robot transfer. This includes demonstrations acquired via kinesthetic teaching or teleoperation.}. Generalization was even termed inter-task transfer learning~\citep{FERNANDEZ2010}. 

Task transfer has also been tackled via meta-learning. In the meta-learning setting, a model is trained on a variety of tasks so that new tasks are solved by using none (zero-shot) or only a limited amount (few-shot) of additional training data~\citep{finn2017model, nichol2018first}. For instance, MAML~\citep{finn2017model} was shown to generalize to new goal velocities and directions in the half cheetah and ant locomotion tasks of the Gymnasium benchmark~\citep{towers_gymnasium_2023} faster than conventional approaches. Other meta-learning approaches tackle the transfer problem from a different perspective by learning loss functions~\citep{Bechtle21:MetaLearningLoss, Bechtle22:MetaLearningLoss}. The meta-learned loss functions generalize to different tasks, thus alleviating the need of designing task-specific losses.

Thus, in a broad sense of Definition \ref{def:TL_robotics}, such approaches already propose solutions for $\lbrace \robot, \domain, \sourcetask \rbrace \to \lbrace \robot, \domain, \targettask \rbrace$, although they were not called transfer learning. Complete skill models were learned from a set of executions also with DNNs~\citep{Loncarevic2022}. The adaptation of the skill model for a new environment is commonly referred to as transfer learning.

\subsection{Robot Transfer}
Above mentioned approaches use knowledge from several instances of a task. However, learning of even one instance of a task could pose a challenge. Imitation learning, where human skill knowledge was transferred to a robot, has been thoroughly researched as the means for learning of task models and their execution on a robot~\citep{Billard2008, Ravichandar2020}. Imitation learning (IL), also known as programming by demonstration (PbD), is in a strict sense an example where the task and the environment remain the same, but the agent is different, $\lbrace \sourcerobot, \domain, \task \rbrace \to \lbrace \targetrobot, \domain, \task \rbrace$, since one of the agents is in fact a person. Note that, in some cases, the environment can also change. In PbD one often transfers the demonstrated motion~\citep{Ijspeert2013}. However, if only the motion is repeated, the task knowledge might be overlooked and the task correspondence~\citep{heyes2001} might not get preserved at all. 
This may be alleviated by transferring other crucial characteristics of the task, so-called task constraints, such as force patterns~\citep{Rozo16:ForceLfD} and posture-dependent task requirements~\citep{Jaquier20:ManipAnalysis}, or by retargeting the demonstrated motion~\citep{Aberman20:SkeletonAwareRetargeting}, e.g., by leveraging optimization methods~\citep{Rakita17:MotionTransfer}, learning approaches, or Riemannian geometry~\citep{Klein2022}.
Task descriptions in the form of reward functions learned from demonstrations are also promising for transferring tasks across different robots. For instance, cross-embodiment inverse reinforcement learning (XIRL)~\citep{Zakka22:XIRL} learns a notion of task progress from demonstrations, which is then used as a reward for robots with different embodiments that successfully learn to reproduce the task.

\section{Challenges and Promising Research Directions}
\label{sec:ChallengesAndPromisingDirections}
The aforementioned examples highlight that knowledge can be transferred across several robots, tasks, and environments, thus highlighting the potential of transfer learning for robotics. However, several key questions falling under the areas of identifying the similarities and differences across tasks, environments, and robot to single out what should be transferred and when remain to be answered to realize the full potential of transfer learning in robotics.
In this section, we describe the key challenges that currently constitute roadblocks on the way to the future of transfer learning in robotics.

\begin{figure*}[tbp]
    \centering
    \includegraphics[width=.85\linewidth]{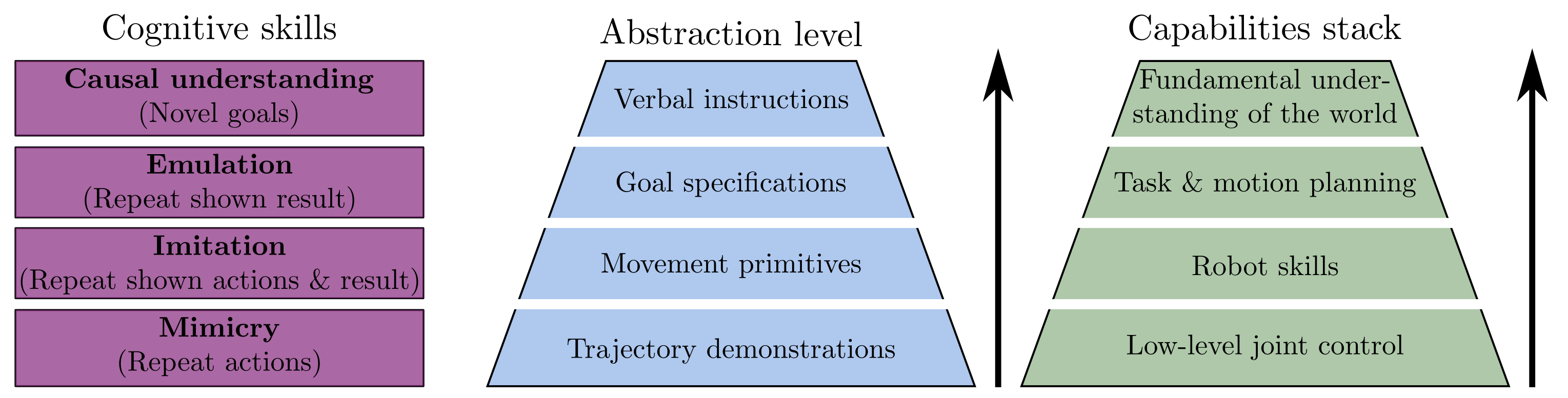}
    \caption{Cognitive skill levels in biology, corresponding abstraction levels in robotics, and associated robotic capability stack. A higher level of abstraction eases the transfer to different agents, environments, and tasks, but requires more and more complex robot's capabilities. Namely, transfer at a given abstraction level requires the robot to be endowed with abilities ranging from the bottom to the corresponding level of the capability stack.}
    \label{fig:AbstractionLevels}
\end{figure*}

\noindent
\subsection{Abstraction Levels in Robotics}
Humans and some animals, such as great apes, acquire cognitive skills via the concept of social learning~\citep{Whiten92:SocialLearning}, whose main component is to copy (transfer) behavior from one individual to another. Social learning takes 
place at different levels depending on the goal and context. In biology, the lowest level of transfer corresponds to \textit{mimicry}, where an individual mimics the actions of another individual superficially, i.e., without any underlying understanding of the goal~\citep{genschow2017mimicry}. Instead, with a number of methodological differences, \textit{imitation} refers to an individual, i.e., the learner, copying the actions of another individual, i.e., the teacher, with the aim of achieving the same goal. As opposed to mimicry, imitation implies an explicit understanding of the goal. At the next level, \textit{emulation} refers to the case where the learner aims at achieving the same goal as the teacher without copying their motor actions~\citep{whiten2004apes}.
Combining imitation, emulation, and some other techniques such as object movement reenactment, the agent ultimately develops an understanding of the world without having to understand the theoretical concept of causality.

These cognitive skill levels can also be roughly identified in robotics, where they intrinsically correspond to different \emph{abstraction levels} (see Figure~\ref{fig:AbstractionLevels}). 
At the lowest learning level, a robot simply mimics the motion of a teacher
without an explicit understanding of the underlying goal. 
If the teacher and the learner have similar embodiments, the task can simply be abstracted as a joint-level (positions, velocity, or acceleration) trajectory. However, in the case of different embodiments, transferring joint trajectories will result in very different end-effector trajectories. 
The task's abstraction level can be increased by specifying, for example, end-effector trajectories and leveraging Cartesian trajectory controllers.
To deal with changes in the environment, both the learning and abstraction levels need to be raised. For instance, transferring end-effector trajectories fails if obstacles are present in the environment. In this case, the task needs to be imitated instead of mimicked, i.e., the goal must be explicitly identified by the robot. The task can therefore be abstracted using, e.g., movement primitives~\citep{Ijspeert02:MPs}, thus allowing the specification of the key components of the imitated trajectory, e.g., the goal position, while leveraging robot skills such as collision avoidance, localization, and object detection to reproduce the task in different environments. In some cases, the physical capabilities of the teacher and the learner are very different, so the learner cannot achieve a demonstrated task by imitating the teacher. Instead, the learner must infer the goal from the demonstrated task and develop a strategy to achieve the same goal~\citep{SCHAAL1999}. In other words, the transfer should be conducted at the higher abstraction level corresponding to achieving the goal specifications without imitating the teacher-specific actions. This corresponds to the emulation learning level. Finally, on an even higher abstraction level, the teacher should ideally give only high-level verbal instruction to the robot such as ``open the drawer", or ``clean the room". This requires the robot to have a skill set resembling that of agents with higher cognitive functions.
Such capabilities have the potential to facilitate transfer in more challenging settings, such as dual- and triple-mode transfer.

To elaborate on the different levels of ``abstraction'', consider the task of transferring a grasp performed by a \emph{source} hand (robot or human) to a \emph{target} robotic hand.
This \emph{transfer} can be performed on three levels: \emph{(1)} The joint angle level~\citep{bouzit1996design, kyriakopoulos1997kinematic} involves directly replicating joint angles with minor adjustments if the hands exhibit similar kinematic structures and degrees of freedom (DoFs); \emph{(ii)} The contact level~\citep{peer2008multi, maeda2016acquiring} is applicable when both hands have an equal number of fingers but differ in their kinematic properties (e.g., DoFs, finger lengths). In this scenario, the target hand strives to grasp the object by replicating the contact positions of the source hand; and \emph{(iii)} The outcome level~\citep{Mahler19:DexterousNet} consists of learning new grasps by optimizing the grasp success with different target hands. 

It is important to notice that the abstraction level has a direct influence on the capabilities that are required for successful transfer across robots, environments, and tasks. In particular, transfer at a given abstraction level requires abilities ranging from the bottom of the robot capability stack onto the abilities of the current level (see Figure~\ref{fig:AbstractionLevels}-\emph{right}). For instance, transfer at the level of verbal instructions demands robots not only to have an abstract understanding of the word, but also to be endowed with task and motion planners, a set of robot skills, and low-level controllers to successfully execute the target task on the target robot in the target environment.
In this context, recent advances in foundational models are a promising research direction to endow robots with emulated high-level cognitive capabilities~\citep{bommasani2021opportunities,Ahn22:SayCan,Driess23:PaLME}. 
Such foundation models generate semantic plans required to execute a target task based on language and on continuous information collected by the robot (e.g., images, state vectors). Driess et al.~\citep{Driess23:PaLME} proposed to address the correspondence problem between tasks at the highest level, i.e., from a semantic perspective, by combining a large language model with perceptual inputs in an embodied multimodal model.
Transfer between robots, tasks, and environments is then achieved via a large amount of training data and by training the models on several robots, tasks, and environments simultaneously. In other words, the transfer comes --- to some extent --- ``for free" thanks to the large scale of foundation models. Importantly, such transfer happens only at the highest level, i.e., at the level of semantic planning, while low-level policies and planners are assumed to be given. 
In other words, transfer is not tackled at lower levels. As a consequence, the difficulty of transfer, as well as the resulting performance, is highly dependent on the capability stack that is made available a priori for each robot. Moreover, training a (still limited) low-level capability stack from scratch, as done, e.g., in~\citep{Ahn22:SayCan}, requires months of data collection and is not scalable in the long run. 

In this sense, we contend that investigating transfer learning methods across the entire robot capability stack is of utmost importance. In particular, we believe that bridging the gap between high-level semantic task transfer~\citep{Driess23:PaLME} and low-level execution of various tasks with different robots in the real world is a crucial challenge for transfer learning in robotics. These require \emph{grounding} the aforementioned transferable high-level representations into the real world via robot sensorimotor experience. 
Such grounded understanding of the world may enable imitation and emulation learning to be intrinsically linked to the robot's physical capabilities, thus facilitating the inference of what can be transferred, at which level, and in which situation. Previous works aiming at grounding language in robot sensorimotor behaviors (see, e.g.,~\citep{KRUGER2011,Cangelosi10:Grounding} may serve as a starting point to tackle this problem. An important challenge is to design grounded representations that allow the expansion of the robot capability stacks at all levels based on similarities between tasks, environments, and robots, thus avoiding cumbersome training of medium- and low-level abilities in novel settings. 
In addition, designing shared grounded representations as proposed in ~\citep{KRUGER2011, Montesano2008} is crucial for transfer across different abstraction levels.

\subsection{Robotics Transformers}

As previously mentioned, the use of large pre-trained foundational models~\citep{bommasani2021opportunities} to learn to transfer is enticing. Several large transformer-based models have been adapted for use in robotics, resulting into so-called \textit{Robotic Transformers}. These models take images and natural language instructions as input and aim to output direct robot actions in the form of Cartesian trajectories. Robotics Transformers were popularized by RT-1~\citep{brohan2022rt}, in which both the input sequence of images and the natural language instructions were tokenized, i.e., broken down into individual units --- words or subwordsfor language and patches for images --- called tokens. 
RT-1 essentially consists of a combination of existing architectures. Namely, the natural Language instructions are first embedded using the universal sentence encoder~\citep{cer2018universal} and passed into a FiLM layer~\citep{perez2018film}, which then constitutes the first layer of EfficientNet-B3~\citep{tan2019efficientnet}, thus allowing the fusion of images and language instructions into tokens. To achieve a closed loop action generation at $3$Hz, the number of tokens is reduced with the TokenLearner~\citep{ryoo2021tokenlearner}. The obtained sequence of tokens, corresponding to the sequence of images, is then finally fed into the transformer architecture~\citep{vaswani2017attention}, which outputs the action consisting $11$ discrete variables of $256$ bins ($7$ variables for the arm and gripper movement, $3$ variables for moving the base, and $1$ variable that switches between controlling the arm, the base, or terminating the episode).
The model is trained with a large dataset of approximately $130000$ episodes performing over $700$ tasks collected in the real world.
Despite the incorporation of semantic reasoning, as well as the considerable amount of training data and model parameters, RT-1 generalization is limited to the combination of seen concepts. Moreover, it is limited to simple robotic tasks, cannot, e.g., generate compliant motions or solve complex and dexterous manipulation tasks, and cannot outperform the task demonstrator.

The subsequent RT-2~\citep{brohan2023rt} is a vision-language-action model based on vision-language models~\citep{chen2023pali,Driess23:PaLME} trained on web-scale data
and tuned with robotic actions. The largest RT-2 consists of $55$ billions parameters.
The increased performance of RT-2 compared to RT-1 and other adjusted baseline models (such as VC-1~\citep{majumdar2023we}, R3M~\citep{nair2022learning}, MOO~\citep{stone2023open}) is attributed to the vision-language backbone combining co-finetuning the pre-trained model jointly on robotics and web data, so that the model considers more abstract visual concepts as well as robot actions. 
Interestingly, the largest RT-2 model displays encouraging emergent capabilities, where the model is able to use the high-level concepts acquired from the web-scale data such as relative relations between objects to complete tasks that were not present in that form in the robotic dataset.
However, these emerging capabilities only emerged in the largest models, which necessitate a complex cloud infrastructure to be deployed. Therefore, they are currently unsuitable for deployment on robotics platforms and self-sufficient autonomous systems.
Moreover, the model is not able to produce motions that are not covered by the large robotics dataset. Furthermore, the size of the model ($55$ billions parameters) can slow the model inference down to $1$Hz. 

Overall, robotics transformers incorporated high-level semantic reasoning capabilities directly into the robotic actions. This is equivalent to fusing the capability stack of Figure~\ref{fig:AbstractionLevels} into a single monolith model. This approach comes at the price of reduced low-level performance and limited capabilities compared to traditional methods that output continuous actions, or direct force control for compliant tasks, among others.
In addition, fusing the capability stack also results in big and cumbersome models, which are difficult to deploy. In this sense, decomposing the capability stack may result in scalable models resulting in higher performances in robotics tasks.

Importantly, robotics transformers still offer potential beyond their usage as an end-to-end controller. Such models can potentially be used in a similar fashion as large pre-trained models have been used in machine learning to obtain compact or universal representations.
In this context, it is worth highlighting RT-X~\citep{padalkar2023open} the latest iteration of the robotic transformers, which aggregated $60$ robotic datasets with $22$ different manipulator embodiments and made this data suitable for the robotic transformer architecture. Such open-source tools and datasets are crucial to bootstrap research in transfer learning for robotics.

\subsection{Universal Representations}
Pretrained representations are widely popular both in machine learning and in robotics~\citep{pari2021surprising}. For instance, ImageNet~\citep{deng2009imagenet} was often used to acquire low-dimensional image representation for picking via suction and parallel gripper~\citep{yen2020learning},  contact-rich high-dimensional
dexterous manipulation tasks~\citep{Kumar21:RRL}, and household tasks such as scooping involving tools~\citep{Liu18:ImitationFromObservation}. 
The main advantage of such representations is their flexibility in being leveraged for many different downstream tasks with little adaptation. 
Although these representations are promising, they still require fine-tuning to be transferred to different settings.
In this sense, robotics would benefit from truly universal representations that would be intrinsically transferable between robots, environments, and complex tasks without additional training. 

In this context, adapting methods such as \emph{universal domain adaptation}~\citep{you2019universal} to robotics stands as a particularly promising research direction. 
Universal domain adaption removes many assumptions regarding the relationship between source and target label dataset. After extracting features from both domains, the proposed universal adaptation network (UAN) employs \emph{(1)} an adversarial discriminator to match the source and target feature distributions falling under common labels, \emph{(2)} a non-adversarial discriminator to obtain the domain similarity, i.e., quantify the similarity of an input with the source domain, and \emph{(3)} a label classifier predicting the probability of the input over to the source classes. Given the domain similarity and the label predicted by the classifier, UAN predicts either a known source label or an \textit{unknown} class label, thus enabling its use in settings where source and target labels are different.
Extending the universal domain adaptation framework beyond classification tasks would be a first promising step towards universal representations for robotics. Such representations may then be directly leveraged for planning and control.

Alternatively, universal representations may be constructed by tasking models with so-called \emph{pretext tasks}, i.e., tasks designed solely to acquire representations that are then used in a plethora of downstream tasks. In unsupervised visual representation learning, the pretext task of instance discrimination~\citep{wu2018unsupervised} inspired many representation models based on contrastive learning~\citep{chen2020simple, he2020momentum, caron2020unsupervised}. Importantly, the pretext task does not require any labels. In other words, the unsupervised setting removes any assumptions on source and target labels, similarly as in universal domain adaptation. Training models unsupervisedly and jointly on source and target data may be a promising direction to obtain universal representations for transfer learning in robotics.

Moreover, the advent of large language models and visual language models brought a new breed of representation models that have been rapidly applied in all areas of robotics, e.g., in planning~\citep{shah2022robotic, huang2022language}, manipulation~\citep{jiang2022vima, ren2023leveraging, khandelwal2022simple}, and navigation~\citep{lin2022adapt, parisi2022unsurprising, gadre2022clip}. 
Such models are excellent candidates to harvest novel universal representations for robotics. 

Some large visual language models jointly account for different modalities by encoding them in a shared latent space. For instance, CLIP~\citep{radford2021learning} is pre-trained on a large dataset of images and associated textual descriptions. The model maps both modalities into a shared latent space using a contrastive loss function. The idea of combining representations from different modalities into a shared latent space was also explored in robotics. For instance,~\citet{Tatiya20:HapticKnowledgeTransfer, Tatiya23:TransferImplicitKnowledge} learned a common latent space from haptic feature space of multiple robots. Knowledge from source robots was then transferred through the latent space to facilitate object recognition by a target robot.
\citet{lee2020making} considered specific encoders for RGB, depth, force-torque, and proprioception modalities, which were aggregated into a multimodal representation with a multimodal fusion model. This shared representation was shown to improve the sample efficiency of the manipulation policy for a peg insertion task. The case of missing modalities during inference time was considered in~\citep{Silva20:PlayingGamesDark}. A perceptual model of the world was trained by assuming that some modalities may not be available at all times. The approach can therefore compensate for missing or corrupted modalities during execution. Such joint representations are crucial to design universal representations for transfer.

To be successfully leveraged in various robotic scenarios, universal representations should be expressive, while remaining simple enough to facilitate downstream applications. This is usually achieved via a dimensionality reduction process by extracting low-dimensional latent representations from data. While this latent space was usually assumed to be Euclidean, i.e., flat, recent works have shown the superiority of curved spaces --- manifolds like hypersphere, hyperbolic spaces, symmetric spaces, and product of thereof --- to learn representations of data exhibiting hierarchical or cyclic structures~\citep{Nickel17:PoincareEmbeddings,Gu18:MixtedCurvatureRepresentations,Lopez21:SymmetricSpacesEmbeddings}. 
For instance, the compositionality of visual scenes can be preserved via hyperbolic latent representations, thus improving downstream performance in point cloud analysis~\citep{Montanaro22:PointcloudsHyperbolic} and unsupervised visual representation learning~\citep{Ge23:HyperbolicContrastiveLearning}. This suggests that rethinking inductive bias in the form of the geometry of universal representations may also be relevant for robotics applications and for transfer learning in robotics. For example, data associated with robotics taxonomies are better represented in hyperbolic spaces~\citep{Jaquier2022:Hyperbolic} and manipulation tasks encoded as graphs in the context of visual action planning~\citep{lippi2022enabling} may benefit from non-Euclidean representations.

\subsection{Interpretability}
Interpretability and explainability of learning-based approaches are key to safely deploy robots into the real world. In particular, black-box approaches lacking human-level interpretability can severely hinder natural and safe interactions with robots. 
In this context, transferable universal representations should also be interpretable and explainable. To do so, approaches in the field of visual action planning~\citep{lippi2022enabling, wang2019learning} proposed to decode the underlying representations into a human-readable format, i.e., images. Alternatively, representations can be readily encoded into a human-readable format that is additionally interpretable by many other methods or software architectures. For instance, the universal scene description (USD)~\citep{Pixar:USD}
was designed to interchange 3D graphics information. This format was recently enhanced by Nvidia to facilitate large, complex digital twins --- reflections of the real world that can be coupled to physical robots and synchronized in real time~\citep{Nvidia:USD}.
USD is made from sets of data structures and APIs, which are then used to represent and modify virtual environments on supported frameworks such as Omniverse~\citep{mittal2023orbit}, Maya~\citep{maya}, and Houdini~\citep{Houdini}. 
Such a framework has significant potential to be used for robotics transferability. For instance, it could be leveraged to build joint representations of the world shared across multiple robots, to share knowledge, and even to infer digital twins from sensory readings.

\subsection{Benchmarking and Simulation}
Benchmarks and relevant metrics are key to evaluate and compare methods, thus having the potential to boost the development of innovative novel approaches. For instance, the rapid improvement of deep-learning models benefited from easily-accessible benchmarks that are widely accepted by the community~\citep{krizhevsky2009learning, deng2009imagenet, lin2014microsoft, cordts2016cityscapes}. 
The robotics community also benefited from impressive strides towards unified benchmarks with efforts such as the YCB-\citep{calli2015ycb} and KIT-\citep{kasper2012kit} object dataset, and with regularly-organized benchmark competitions such as RoboCup~\citep{kitano1997robocup}, ANA Avatar Xprice\footnote{\url{https://www.xprize.org/prizes/avatar}}, and DARPA challenges\footnote{\url{https://www.darpa.mil/program/darpa-robotics-challenge}}. However, they all face robotics unique challenges.
First, as robots are real systems evolving in the real world, the deployment of any method can be highly time-consuming. Second, as previously mentioned, transferring methods to robots with different embodiments is non-trivial, which intrinsically hinders benchmarking across different research groups. Last but least, handcrafted, highly tuned solutions usually outperform more general methods to solve any specific or standardized task as defined in classical benchmarks. This is especially notable for robotic manipulation where accepted benchmarks remain scarce.

The Robothon 2023 task board challenge~\citep{so2022towards} is an example of recent robotics manipulation benchmark.
This board is an assembly of various relevant robotics tasks --- including inserting a key into a keyhole and turning it, plugging/unplugging an ethernet connector, and pushing switches, among others --- allowing the evaluation of different approaches.
As required for a benchmark, the task board is standardized and its specifications are given. However, in such settings, handcrafted, or even prerecorded, motions can lead to surprisingly high scores. Randomly orienting the board before every trial was later included to discourage such solutions. 
Within machine learning benchmarks, handcrafted solutions are prevented by dividing the available data into training and test sets, which consists of different samples drawn from a single distribution.
Analogously, the Robothon 2023 challenge would require a large number of task boards consisting of the same high-level tasks but differing in their geometric-specific realization. A promising avenue to overcome the impracticability of producing numerous physical task boards would be to leverage simulators. 

Modern robotic simulators, such as Mujoco~\citep{todorov2012mujoco}, Bullet~\citep{coumans2021}, and PhysX\footnote{\url{https://developer.nvidia.com/physx-sdk}} have shown impressive improvements in various areas, including in robotics assembly~\citep{narang2022factory}. Such simulators have the potential to generate various parametrizations of simulated boards, and thus to create training and test sets similar to machine learning benchmarks. In particular, such sets would be of high relevance for transferability in robotics, as they have the potential to evaluate transferability across \emph{(1)} robots, \emph{(2)} environments, i.e., different parametrizations, and \emph{(3)} tasks performed on the same board. 
The ultimate challenge is to overcome the sim-to-real gap when deploying the developed methods on a real, previously unknown task board using a new robot during live competition.
Such benchmarks would boost research in transferability in robotics, as well as provide valuable information on the difficulty and challenges of each transfer setting. 
It is worth highlighting that a wide range of works and methods have been developed in the field of machine learning in recent years and subsequently compiled into transfer-learning-libraries~\citep{tllib}. Such a consortium of methods offers a huge potential to be used in robotics contexts. Importantly, relevant metrics must be defined to compare different approaches in different transfer settings.

\subsection{Metrics for Transfer Learning in Robotics}
Two types of metrics are relevant for transfer learning in robotics, namely \emph{(i)} metrics measuring the \emph{quality of the transfer}, and \emph{(ii)} metrics measuring the \emph{transfer gap}. 
Metrics measuring the transfer quality aim at quantifying algorithmic performance and allows the comparison of the performances of different algorithms on the target space after transfer. In particular, they can also determine \textit{when} transfer learning is useful. Metrics measuring the transfer gap measure the the discrepancy between the source and target spaces. Essentially, they provide a notion of how different the source and target robots, environments, and tasks are.

\textbf{Transfer quality metrics} are crucial to evaluate and compare the performance of transfer learning algorithms. As such, they are key to the development of novel transfer learning methods. 
Transfer quality in machine learning is commonly measured by comparing the performance achieved on the target task with or without transfer learning. 
In this case, the quality metric includes not only the final performance (asymptotic performance), but also the initial benefit of the transfer (jumpstart performance), the time to reach a predefined performance threshold (time to threshold), as well as the sensitivity to different hyperparameter settings~\citep{taylor2009transfer, taylor2007transfer}. In addition to measuring the transfer quality, further analysis of the transfer process can be conducted by comparing the number of required sub-source tasks, the number of demonstrations~\citep{barreto2018transfer, zhu2020off}, or the required quality (e.g., suboptimal, expert, oracle) of the source space~\citep{zhu2020learning}.
Recently, \citet{Chen23:EvaluationMetricsUDA} highlighted the importance of robust unsupervised evaluation metrics for domain adaptation. Such metrics should be independent of the training method,  consistent across hyperparameters and models, and robust to adversarial attacks.
Most of the aforementioned transfer quality metrics can and are in fact already used for transfer learning in robotics~\citep{zhu2023transfer}.

\textbf{Transfer gap metrics} provide a measure of the discrepency between the source and target spaces. Note that such metrics may also be used to measure performance in certain circumstances.
Robotics adds an additional challenge to the problem of defining suitable transfer gap metrics for transfer learning: Indeed, transfer learning in robotics can be seen as a three-part transfer problem consisting of transfer across robots, tasks, and environments. 
Several metrics have been defined and directly optimized to solve each of these sub-problems.   
In this context, domain adaptation received considerable attention from the machine learning community in recent years. When the distribution of the source and target domains can be reliably estimated, simple divergences, e.g., the Kullback-Leibler (KL) divergence~\citep{kullback1951information} the Maximum Mean Discrepancy (MMD)~\citep{gretton2012kernel} for labeled data, or the $H\Delta H$ divergence~\citep{ben2010theory}
MDD~\citep{zhang2019bridging}, and SND~\citep{saito2021tune} for unlabeled data, provide a quantitative estimate of the domain transfer gap.
In robotics, a large body of work focuses on the sim-to-real gap --- or in other words, the reality gap --- as a specific domain gap $G_{\domain}$.
The sim-to-real gap is often measured as the capacity of a \textit{realistic} simulator to emulate the real world. Collins et al.~\citep{collins2019quantifying} quantified the reality gap by comparing simulated robot trajectories, 
e.g. using Pybullet~\citep{coumans2016pybullet} or Mujoco~\citep{todorov2012mujoco}, with real-world trajectories captured by a motion capturing system.
Importantly, the simulators accurately model kinematics, but generally struggle with dynamics of robots interacting with objects. Zhang et al.~\citep{zhang2020predicting} specifically focused on the sim-to-real gap in robotics and predicted the transfer performance of reinforcement learning policies using a probabilistic dynamics model.
Limited attention was devoted to designing transfer gap metrics for transfer learning across tasks or robots. In particular, the existing literature related to skill transfer learning in human-robot cooperation~\citep{liu2020skill} does not agree on a specific skill transfer metric to measure the task transfer gap $G_{\task}$. Although various metrics have also been proposed in the context of motion retargeting, see, e.g.,~\citep{Gielniak13:HumanLikeMotionGeneration,Penco18:MotionRetargeting}, quantifying the quality of retargeted motions, as well as robot transfer gap $G_{\robot}$, generally remain open questions.
We contend that a suitable transfer gap metric $G$ for robotics should consider all three settings of transfer. For instance, this metric may be defined as a simple combination of individual metrics for robot, environment, and task transfer, e.g., 
\begin{equation*}
    G = \lambda_1 G_{\robot} + \lambda_2 G_{\domain} + \lambda_3 G_{\task},
\end{equation*}
where $\lambda_1, \lambda_2, \lambda_3$ are weights adjusting the individual metric influence.
Such transfer metric has the potential to bootstrap the development of different transfer learning methods for robotics, as it gives insights about the discrepancy of each mode (i.e., robot, environment, and tasks) from source to target space.

\subsection{Negative Transfer} 
Importantly, transfer learning is not necessarily beneficial in all settings.
Transfer learning algorithms build on systematic similarities between source and target spaces. However, if non-existing similarities are selected by the algorithm, the transfer can have a negative impact on the performance in the target space~\citep{wang2021mitigating}. This phenomena is denoted \emph{negative transfer}~\citep{rosenstein2005transfer}. 
Negative transfer has notably been studied within the field of meta-learning~\citep{thrun1998learning}, in which a rapid adaptation to the novel task is assumed to be key for the success of the corresponding models. 
Preliminary work~\citep{deleu2018effects} showed that adaptation using meta-learning algorithms, such model-agnostic meta-learning (MAML)~\citep{finn2017model}, can significantly reduce the performance on meta-training tasks.

In robotics, negative transfer may occur at the different levels of the robot capability stack. For instance, at the low control level, transferring an inverse dynamic model learned for a source quadrotor to a target quadrotor with significantly different physical properties has been shown to lead to worse performances than using a baseline controller that disregards the inverse dynamics~\citep{Sorocky20:MultiSourceTransfer,Sorocky21:ShareOrNot}. 
Interestingly, the lower levels of the capability stack may be more susceptible to negative transfer as low-level information, e.g., inverse dynamic models, may only be transferred across closely-related source and target spaces. In contrast, experience at the higher levels is more general and may be transferred across a larger range of source and target spaces.

We believe that negative transfer remains an under-investigated direction in robotics. In particular, negative transfer may be particularly harmful for robotics. First, negative transfer may lead to potentially-damaging behaviors of the target robot, while safety is a crucial aspect when deploying robots in the real world. Second, negative transfer may lead to transfer learning requiring longer training time than directly learning the desired behavior in the target space, while low training time is crucial for real robots acting in the real world. Therefore, the effects and causes of negative learning remain to be thoroughly studied, as they may be key to develop successful and reliable transfer learning algorithms tailored to robotics.

\section{Conclusion: The Future of Transfer Learning in Robotics}
\label{sec:Conclusion}
The rise of transfer learning implies its potential to enable robots to leverage available knowledge to learn and master novel situations efficiently. In this paper, we aimed at unifying the concept of transfer learning in robotics via a novel taxonomy acting as a bedrock for future developments in the field. Building on the successes of transfer learning in robotics, we outlined relevant challenges that have to be solved to realize its full potential. 
It is important to highlight that these challenges intrinsically relate to determining what can and cannot be transferred. As illustrated in this paper, transfer learning in machine learning largely relies on identifying whether the distribution of data across two domains remain similar. Identifying similarities and differences across robotic tasks amounts to more than comparing distributions. We have emphasized the need to delineate \emph{similarities across tasks, environments, and robots} in an effort to ease identification of commonalities and differences in each case. 
Automatically identify similarities across two situations in robotics relies importantly on spelling out how much prior knowledge on the physics of the robot and environment is provided. With advances in the use of foundational models, the amount of prior knowledge readily available may ease this transfer.

We hope that this position paper paves the way towards successful transfer learning between robots, tasks, and environments, as well as their compositions. Reusing knowledge holds the promise of closing the performance gap between humans and robots in overcoming novel challenges and acquiring new skills and concepts.

\begin{acks}
The research leading to these results has received funding from the European Union's Horizon Europe Framework Programme under grant agreement No 101070596 (euROBIN).
\end{acks}

%\clearpage
\bibliography{bibliography}
\bibliographystyle{SageH}

\end{document}